\pdfoutput=1
\documentclass[%
    11pt
]{article}
\usepackage[]{acl}

\usepackage{latexsym}
\usepackage[T1]{fontenc}
\usepackage[utf8]{inputenc}
\usepackage{newtxtext,newtxmath}
\usepackage{inconsolata}
\usepackage{microtype}
\usepackage{nicefrac}
\usepackage{siunitx}
\usepackage{graphicx}
\usepackage{csquotes}
\usepackage{tabularx}
\usepackage{booktabs}
\usepackage{multirow}
\usepackage{hyperref}
\usepackage{balance}
\usepackage{subcaption}
\usepackage{newfloat}
\DeclareFloatingEnvironment[
  fileext=ex,
  listname={List of Examples},
  name=Example
]{example}
\DeclareCaptionSubType{example}

\title{Are Multimodal Large Language Models Pragmatically Competent Listeners in Simple Reference Resolution Tasks?}

\author{
    Simeon Junker \: 
    Manar Ali \:
    Larissa Koch \:
    Sina Zarrieß \:
    Hendrik Buschmeier
\\
    CRC 1646 ‘Linguistic Creativity in Communication’
\\ 
    Bielefeld University, Bielefeld, Germany
}

\begin{document}

\maketitle
\begin{abstract}
    We investigate the linguistic abilities of \emph{multimodal} large language models in reference resolution tasks featuring simple yet abstract visual stimuli, such as color patches and color grids. Although the task may not seem challenging for today's language models, being straightforward for human dyads, we consider it to be a highly relevant probe of the pragmatic capabilities of MLLMs. Our results and analyses indeed suggest that basic pragmatic capabilities, such as context-dependent interpretation of color descriptions, still constitute major challenges for state-of-the-art MLLMs.
\end{abstract}

\section{Introduction}
\label{sec:introduction}

The advent of large language models (LLMs) and their expansion in scale, variety, and availability over the past decade has led to considerable interest -- both in the research community as well as in the public at large -- in understanding their capabilities and limitations. A lot of research focuses on their general abilities (reasoning, math, professional examinations, …) in order to answer questions about the level or category of ‘intelligence’ these models exhibit \citep[e.g.,][]{BubeckChandrasekaran2023} or which practical tasks they might be suitable for. In contrast to this, research in (computational) linguistics is also interested in their basic linguistic abilities \citep{chang-bergen-2024-language, milliere2024language}.

\begin{figure}
    
    \begin{subfigure}{\columnwidth}
        \centering
        \includegraphics[width=\columnwidth]{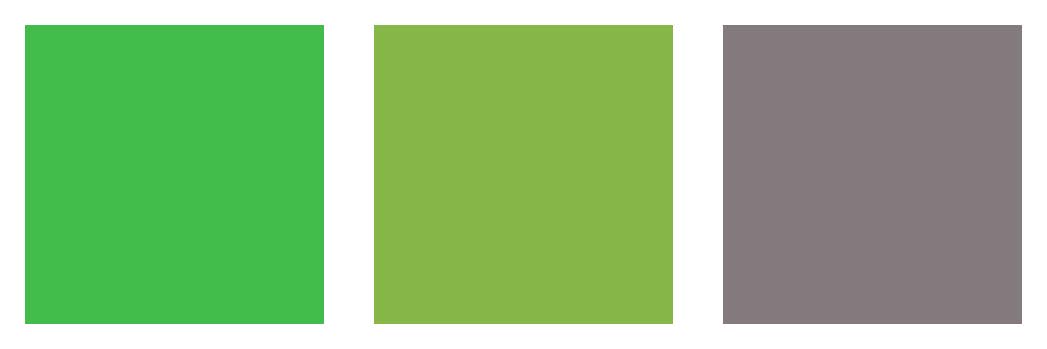}
        \footnotesize
        \begin{tabularx}{\columnwidth}{lX}
            director: & \emph{yellow green. the less green one} \\
            model responses: & $4 \times$ left, $4 \times $ \textbf{middle}, $0 \times $ right
        \end{tabularx}    
        \caption{Color patch example \citep{monroe-etal-2017-colors}.}
        \label{fig:patches}
        \vspace{1mm}
    \end{subfigure}
    
    \begin{subfigure}{\columnwidth}
        \centering        
        \includegraphics[width=\columnwidth]{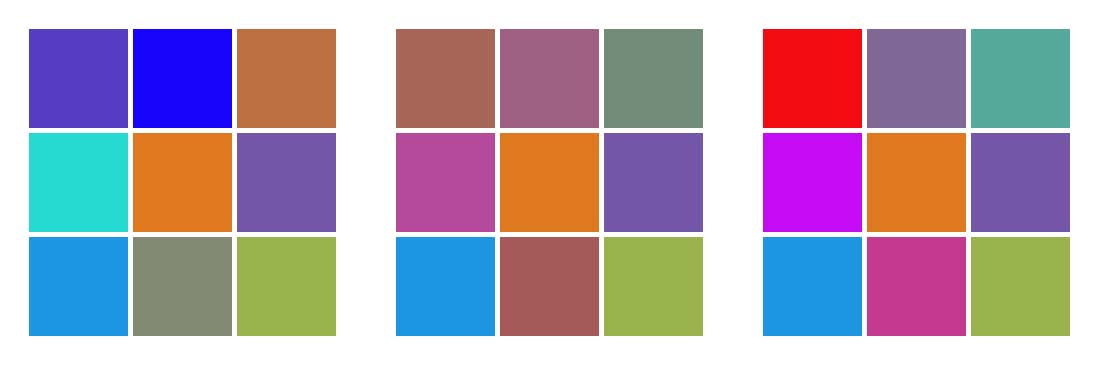}
        \footnotesize
        \begin{tabularx}{\columnwidth}{lX}
            director: & \emph{first square is brown}\\
            model responses: & $5 \times$ left, $3 \times $ \textbf{middle}, $0 \times $ right
        \end{tabularx}
        \caption{Color grid example \citep{mcdowell-goodman-2019-learning}.}
        \label{fig:grids}
    \end{subfigure}

\caption{%
    Example color patches and color grids stimuli for director-matcher-style dyadic reference games with the human director's description and reference resolution responses of eight different MLLMs. In both examples the target referent is the object in the middle and was correctly identified by the human matcher.}
\label{fig:examples}
\end{figure}

In this paper, we investigate the linguistic abilities of \emph{multimodal} large language models (MLLMs) on the level of language use, i.e., linguistic pragmatics, in the well-known reference resolution paradigm. More specifically, we examine whether off-the-shelf MLLMs (LLaVA-NeXT, Qwen2-VL, and Janus-Pro) are able to resolve references to abstract visual stimuli (color patches and color grids; \citealp{monroe-etal-2017-colors, mcdowell-goodman-2019-learning}; see Fig.~\ref{fig:examples}) given in director-matcher-style dyadic reference games \citep{ClarkWilkes-Gibbs1986}.

Although the task may not seem challenging for today's language models because it is straightforward and easy for human dyads, we consider it to be a highly relevant probe of the pragmatic capabilities of MLLMs. Being ubiquitous in all of language use, reference (generation and resolution) is highly context-dependent (here visual context). The same can be said about the color references at the center of the task (where color is a main distinguishing attribute between potential referents; \citealp{monroe-etal-2017-colors}). The complexity of the visual context varies between the two types of stimuli we consider, but can generally be considered simple. Their abstractness, however, poses demands on the basic visual perception capabilities of MLLMs.

Thus, we investigate how well MLLMs perform basic reference resolution tasks requiring contextualized pragmatic reasoning about color and simple spatial arrangements in two abstract visual domains.\footnote{%
    Code and data of the study are available at 
    \href{https://doi.org/10.5281/zenodo.15553655}{https://doi.org/ 10.5281/zenodo.15553655} as well as
    \href{https://github.com/clause-bielefeld/mllm-listeners}{https://github.com/clause-bielefeld/mllm-listeners}.}
Our results show that models with sufficient capacity achieve promising results for color patches. However, even the best-performing models struggle with the complex structure of color grids.

\section{Background}
\label{sec:background}

The semantics and pragmatics of references to color have long played a special role in work on reference games \citep{pechmann1989incremental, baumgaertner-etal-2012-towards, koolen2013effect, zarriess-schlangen-2016-towards}. 
The categorization of colors is well-known to be subject to complex interactions between semantic and perceptual information \citep{mitterer2008recalibrating}.
The naming of colors in interactive games is also well-known to be subject to pragmatic reasoning and negotiation between interaction partners \citep{meo2014generating, mcmahan-stone-2015-bayesian, monroe-etal-2017-colors}.
Finally, color references have been studied from the perspective of figurative language and creativity \citep{kawakami-etal-2016-character}.

Color perception and understanding tasks have also been used in recent work on probing language models.
\citet{loyola-etal-2023-perceptual} explored the alignment between the perceptual color spaces of humans and text-based LLMs. Their findings indicate moderate alignment for basic color terms, which decreased as color descriptions become complex and subjective. Similarly, \citet{abdou-etal-2021-language} found that alignment improves with model size.
\citet{jones-etal-2024-multimodal} examined the sensitivity of MLLMs to sensorimotor features by testing their ability to identify images that match textual implied features, showing that the effect for color emerges only in the largest model.
\citet{Rahmanzadehgervi_2024_ACCV}'s study on vision language models shows that even state-of-the-art models, such as GPT4-o and Gemini-1.5 Pro, perform surprisingly poorly at a range of low-level tasks, which humans are expected to complete with ease, e.g., determining whether two circles overlap.

Together, these studies warrant further investigations and analyses into multimodal language understanding tasks. In this paper, we investigate tasks for analyzing low-level language grounding of color stimuli in combination with pragmatic capabilities, focusing on referring expressions produced by human players of reference games.

\section{Experimental Setting}
\label{sec:experiments}

\subsection{Models}
\label{sec:models}

We investigate reference resolution in the following  MLLMs:
\textbf{LLaVA-NeXT} \cite{liu2024llavanext} builds on
LLaVA 1.5 \cite{Liu_2024_CVPR} and the original LLaVA \cite{LiuLi2023}, with CLIP-ViT-L as the vision encoder. For the LLM backbone, we use Vicuna at the 3b and 7b parameter scales, NousHermes2-Yi at the 34b scale, and Llama 3 at the 72b scale.
\textbf{Qwen2-VL} \citep{Qwen2VL} upgrades the Qwen-VL \citep{Qwen-VL} models and uses ViT as a vision encoder and multimodal rotational position encoding (M-RoPE). It comes with Qwen2 as the LLM backbone and is available with 2b, 7b, and 72b parameters. Qwen2-VL-72B has been reported to perform similarly to state-of-the-art models such as GPT-4o \citep{Qwen2VL}.
\textbf{Janus-Pro} \cite{chen2025janus} enhances the original Janus \cite{wu2024janus}, both of which utilize the structure of decoupling visual encoding (SigLIP for understanding tasks, VQ tokenizer for generation) with an auto-regressive transformer LLM backbone. 
We use it at 1b and 7b parameter scales.
See Table~\ref{tab:model} (Appendix) for a more detailed model overview.

\begin{table*}
    \small
    \begin{tabularx}{\textwidth}{lXXXXXXXXXX}
    \toprule
     &  &  & \multicolumn{4}{c}{\textbf{Color Patches}} & \multicolumn{4}{c}{\textbf{Color Grids}} \\
     \cmidrule(lr){4-7}
     \cmidrule(lr){8-11}
    \textbf{Model} & \textbf{Size (b)} & \textbf{Quant} & \textbf{Total} & \textbf{Far} & \textbf{Split} & \textbf{Close} & \textbf{Total} & \textbf{Far} & \textbf{Split} & \textbf{Close} \\
    \midrule
    \multirow[c]{2}{*}{Janus} & 1 & --- & 36.1 & 38.1 & 35.4 & 34.8 & 33.3 & 33.2 & 33.4 & 33.3 \\
     & 7 & --- & 68.4 & 83.9 & 64.4 & 56.8 & 39.5 & 41.1 & 38.7 & 38.7 \\
     \midrule
    \multirow[c]{4}{*}{LLaVA} & 7 & --- & 60.1 & 71.4 & 59.1 & 49.5 & 38.0 & 38.6 & 38.7 & 36.8 \\
     & 13 & --- & 59.4 & 70.0 & 57.8 & 50.4 & 37.7 & 38.3 & 37.9 & 36.8 \\
     & 34 & --- & 80.3 & 93.1 & 77.4 & 70.2 & 37.9 & 38.8 & 38.2 & 36.6 \\
     & 72 & 8bit & 62.3 & 75.8 & 59.9 & 51.2 & 39.9 & 42.0 & 40.2 & 37.6 \\
     \midrule
    \multirow[c]{3}{*}{Qwen} & 2 & --- & 61.9 & 77.3 & 58.3 & 50.0 & 38.4 & 40.0 & 38.3 & 36.8 \\
     & 7 & --- & 83.0 & 94.1 & 81.1 & 73.8 & 45.2 & 47.7 & 44.4 & 43.4 \\
     & 72 & awq & \textbf{87.5} & \textbf{95.1} & \textbf{86.9} & \textbf{80.3} & \textbf{66.5} & \textbf{70.2} & \textbf{66.0} & \textbf{63.2} \\
     \midrule
    human & --- & --- & 90.0 & 97.0 & 89.7 & 83.3 & 92.7 & 96.0 & 92.4 & 89.8 \\
    \bottomrule
\end{tabularx}
    \caption{Accuracy scores (\%) for all models and datasets. The highest score per column is highlighted in bold.}
    \label{tab:results}
\end{table*}

\subsection{Data}
\label{sec:data}

We base our investigation on two human director-matcher-style reference game data sets, in which a director (speaker) describes a target to a matcher (listener) who must identify the described target (see Fig.~\ref{fig:examples}). Each data point -- one round of a game -- consists of an abstract visual stimulus (e.g., three color patches in random order) and the textual utterances produced by the two participants. In both datasets, the stimuli were created with three degrees of visual complexity (‘far’, ‘split’, and ‘close’), based on CIELAB color distances.

\paragraph{Color Patches} The data set consists of 948 games of 50 rounds each \citep{monroe-etal-2017-colors}. Participants were assigned the role of either ‘director’ or ‘matcher’. In each round of the game, both participants were presented with three color patches (Fig.~\ref{fig:patches}) shown in a different order. The director knows the ‘target’ color patch and has to describe it by text input, for the matcher to identify it. The matcher can clarify and has to select the target. With three colors being displayed, the description is thus not only shaped by the target color itself, but also by its context, the two ‘distractor’ colors from which it must be distinguished.

\paragraph{Color Grids}  The data set consists of 197 games of 60 rounds each \citep{mcdowell-goodman-2019-learning}. It follows the same procedure, but the stimuli consist of three $3\times3$ grids of color patches (Fig.~\ref{fig:grids}).

\subsection{Prompting Procedure}
\label{sec:prompting}

For generating multimodal prompts for the MLLMs, we render color patches and grid items in the same order as they were presented to the human matcher in the original data. We concatenate the three color patches or the three color grid items into a single image before feeding this combined image to the model. The verbal part of the prompt contains the full, original dialogue between director and matcher on a visual stimulus, transformed into a script-like format (\emph{speaker: … \emph{\P} listener: … \emph{\P} …}). We prompt the model to locate the target referent described by the speaker in the dialogue (see Appendix~\ref{app:prompts}), generate the model response using greedy decoding, and extract the location information (i.e., left, middle, or right) from the model response using a regular expression.

\section{Results}
\label{sec:results}

We evaluate the MLLMs' performance with accuracy, i.e., how often the models identify the visual target correctly. We compare results between 
    (a) visual complexity condition (far/close/split, see  Section~\ref{sec:data}) 
    (b) humans and models; and 
    (c) different models and model sizes (see Section~\ref{sec:models}).
Table~\ref{tab:results} shows the main results for color patches and color grids.
In addition, we test our models on restricted subsets of the color patch and grid datasets, where we included only dialogues with a total length of $\leq 5$ words, out of which $\geq \nicefrac{1}{3}$ have to be basic color terms (see Appendix~\ref{app:results-simplified-dialogues}). We applied this restriction to reduce the linguistic complexity and to remove the meta-commentary that often appears in longer exchanges \citep{monroe-etal-2017-colors}.

\subsection{Color Patches}
\label{sec:color-patches}

Accuracies in the color patch task show a wide range of performance levels, across models and model sizes. The smallest Janus model barely performs above chance level (36 \% accuracy), displaying a strong bias to predict \emph{left} positions for targets (see Appendix~\ref{app:location-biases}). 
In contrast, the largest Qwen model comes close to human accuracy (cf. Table \ref{tab:results}).
Qwen 7b also outperforms Janus and LLaVA variants of the same size, and even Qwen 2b achieves competitive results. 
Janus and Qwen generally improve with larger model sizes. For LLaVA, this is much less consistent, likely because the larger models use different LLM backbones (see Section~\ref{sec:models}, Table~\ref{tab:model}, Appendix). Here, the 7b, 13b, and 72b variants achieve very similar scores, with the largest variant outperforming the smallest model only by a 2-point increase in accuracy. LLaVA 34b behaves as an outlier and achieves notable improvements over all other LLaVA variants.

Model performance also depends substantially on the degree of visual complexity, with most models achieving the highest scores in the \emph{far} condition, and the lowest accuracies for the more challenging \emph{close} items. While increasing size in the LLaVA models (except 34b) does not incur any improvement on the \emph{close} condition, we find a substantial improvement for Qwen 72b over 7b on the \emph{close} condition.

On the subset of items with limited description complexity, all models show only small improvements (cf. Table~\ref{tab:simplified_results}, Appendix~\ref{app:results-simplified-dialogues}). Reducing linguistic complexity in the dialogues does not make it substantially easier for the models to resolve the visual targets. This suggests that models face challenges in linguistically simple expressions consisting of a few basic color terms only. Our inspection of examples (Section~\ref{sec:examples}) supports this.
Overall, the results show near-human scores from some models, but also demonstrate that even supposedly simple color patches can pose problems for current MLLMs.

\subsection{Color Grids}
\label{sec:color-grids}

Accuracy scores for color grids (Table~\ref{tab:results}) are considerably lower than for color patches, with the best-performing model (Qwen 72b) falling short of the total human accuracy by more than 25 points.
The lowest performing model (Janus 1b) does not even perform above chance level. 
We again observe that larger models achieve higher accuracies, but only Qwen 72b achieves a somewhat satisfactory accuracy of 66\%, surpassing Qwen 7b by more than 20 points. 
This indicates that all models did not learn certain aspects of multimodal language grounding required in this task, failing to reason about the more challenging abstract grid shapes.

Overall, differences between the visual complexity conditions are less pronounced than for color patches, supporting the impression that models generally struggle with grounding references in the grid shapes. 
Notably, reducing linguistic complexity leads to further decreases in accuracies for the \emph{far} and \emph{split} color conditions (Table~\ref{tab:simplified_results}), indicating that this task exhibits pragmatic complexities beyond simple color understanding.

Part of this is the more complex spatial arrangement of colors in the grid, resulting in frequent spatial descriptions in the dialogues. 
Qualitative inspections of examples led us to suspect that models adopt a simple strategy of spotting spatial keywords and using these in their responses.
To test this, we partition the grid data into smaller sets of dialogues containing exactly one of the position labels “left”, “middle” or “right”, and calculate the frequency in which our models generate the respective label in the responses. 
The results in Table~\ref{tab:loc_bias} (Appendix~\ref{app:location-biases}) show that all of our models are biased towards predicting labels which are also mentioned in the dialogues.
This indicates that all tested models struggle with this nested structure of grids and do not differentiate enough between position descriptions within grids or between potential referents.

\begin{example*}
\centering
    \begin{subexample}{0.3\textwidth}
        \includegraphics[width=\textwidth]{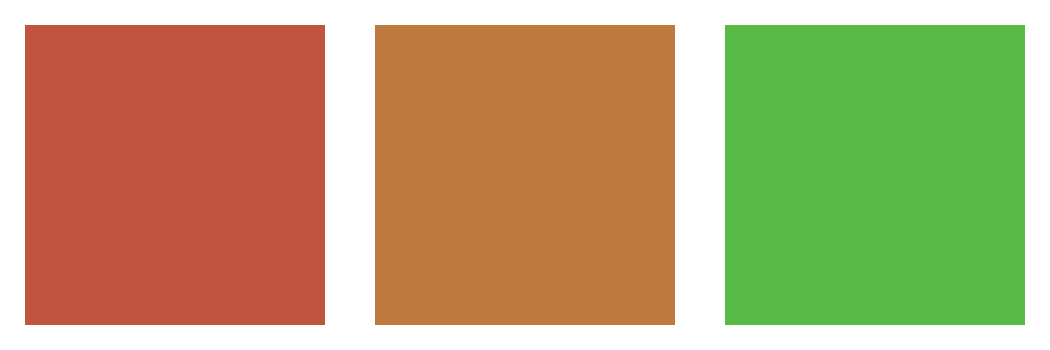}
        \caption{\enquote{orange} (middle)}
        \label{ex:patch_example11}
    \end{subexample}
    \hfill
    \begin{subexample}{0.3\textwidth}
        \includegraphics[width=\textwidth]{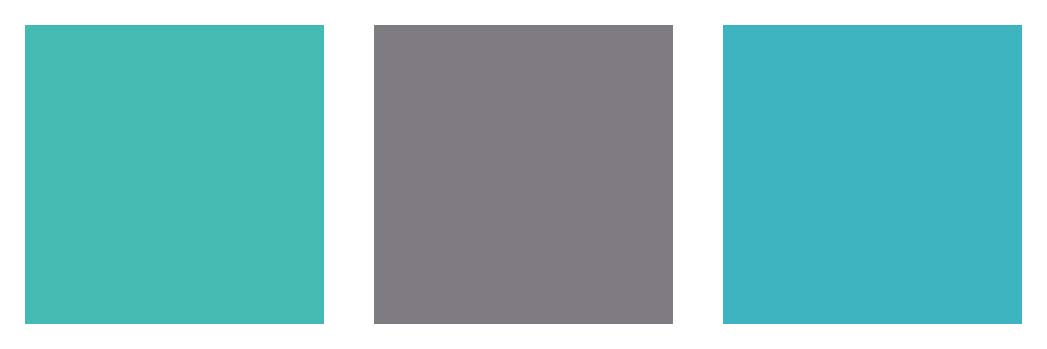}
        \caption{\enquote{more turquoise than cyan} (right)}
        \label{ex:patch_example12}
    \end{subexample}
    \hfill
    \begin{subexample}{0.3\textwidth}
        \includegraphics[width=\textwidth]{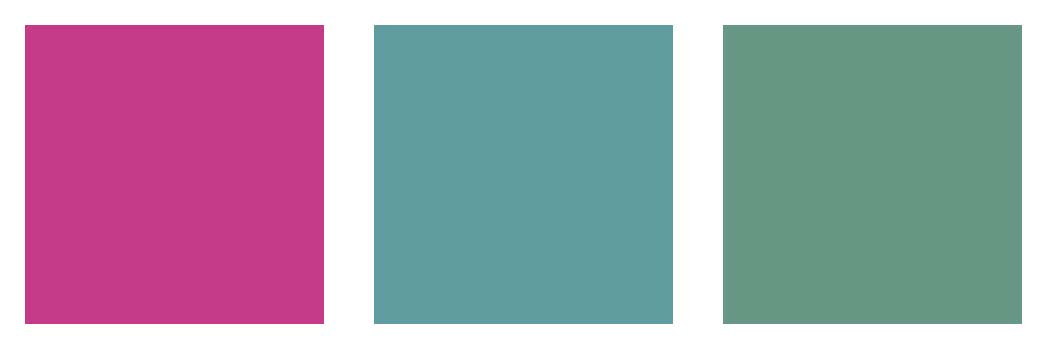}
        \caption{\enquote{bluish, not greenish} (middle)}
        \label{ex:patch_example9}
    \end{subexample}
    \\
    \vspace{2mm} 
    \begin{subexample}{0.48\textwidth}
        \includegraphics[width=\textwidth]{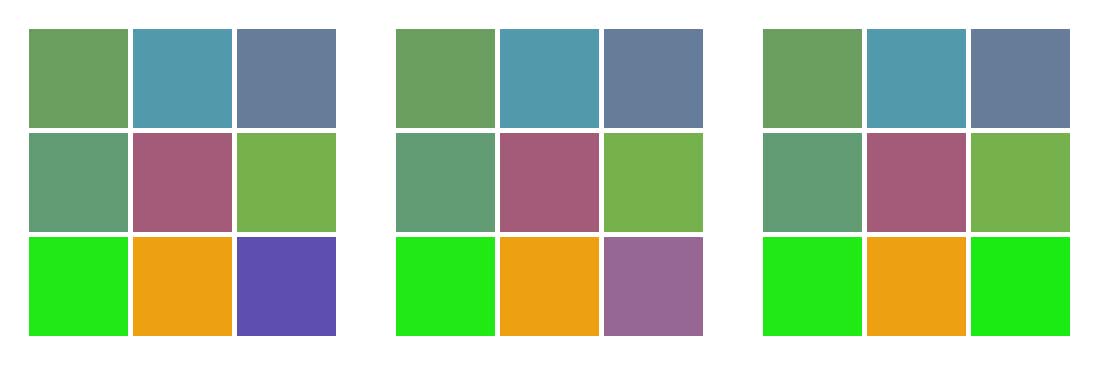}
        \caption{\enquote{the right bottom is a bright purplish blue} (left)}
        \label{ex:grid_example3}
    \end{subexample}
    \hfill
    \begin{subexample}{0.48\textwidth}
        \includegraphics[width=\textwidth]{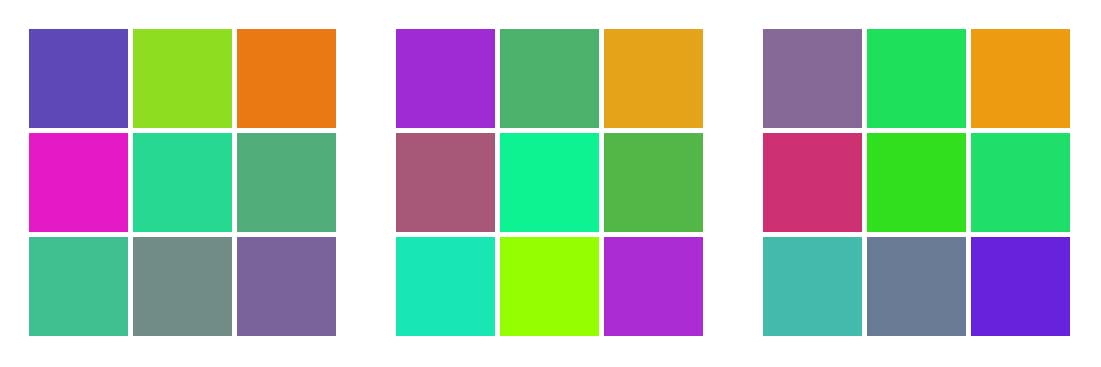}
        \caption{\enquote{middle tile green (not aqua), lower right purple} (right)}
        \label{ex:grid_example14}
    \end{subexample}
    \caption{Examples for color patches (a--c) and color grids (d, e). The caption of each example shows the description of the target given by the director, and the target (left, middle, right). See Table~\ref{tab:examples} for human and model responses.}
    \vspace{3mm}
    \label{ex:examples}
\end{example*}

\begin{table*}
    \small
    \begin{tabularx}{\textwidth}{llllXXXXXXXX}
\toprule
    &&&& \multicolumn{2}{c}{\textbf{Janus}} & \multicolumn{4}{c}{\textbf{LLaVA}} & \multicolumn{2}{c}{\textbf{Qwen}} \\
\cmidrule(l){5-6} \cmidrule(l){7-10} \cmidrule(l){11-12}
    \textbf{Ex.} & \textbf{Condition} & \textbf{Correct} & \textbf{Human} & \textbf{1b} & \textbf{7b} & \textbf{7b} & \textbf{13b} & \textbf{34b} & \textbf{72b} & \textbf{2b} & \textbf{72b} \\
\midrule
    \ref{ex:patch_example11} & split & middle & \textbf{middle} & left & \textbf{middle} & left & left & left & left & \textbf{middle} & \textbf{middle} \\
    \ref{ex:patch_example12} & split & right & \textbf{right} &  left &  left &  left &  left &  left &  left &  left &  left \\
    \ref{ex:patch_example9} & split & middle & \textbf{middle} &  left & \textbf{middle} &  right &  right & \textbf{middle} &  right &  left & \textbf{middle} \\
\midrule
    \ref{ex:grid_example3} & split & left & \textbf{left} &  right &  right &  right &  right &  right &  right &  right &  right \\
    \ref{ex:grid_example14} & close & right & \textbf{right} &  left &  middle &  middle &  middle &  middle &  middle &  middle &  middle \\
\bottomrule
\end{tabularx}

    \caption{Human reference resolution and model responses for the target color patch examples~\ref{ex:patch_example11}–\ref{ex:patch_example9} and color grid examples~\ref{ex:grid_example3} and \ref{ex:grid_example14}. Correctly resolved references are highlighted in boldface.}
    \vspace{1mm}
    \label{tab:examples}
\end{table*}

\subsection{Examples}
\label{sec:examples}

A qualitative inspection of reference resolution failures in the color patch data (Examples~\ref{ex:patch_example11}--\ref{ex:patch_example9} and Table~\ref{tab:examples}) suggests that models struggle with different semantic and pragmatic phenomena. 
In Example~\ref{ex:patch_example11}, the basic color term \enquote{orange} also applies to a distractor, which, however, would rather be called \enquote{red} in this context. Some models show problems with this ambiguity, including all LLaVA variants.
Similarly, in Example~\ref{ex:patch_example12}, all models struggle with a basic but highly context-dependent comparative color description.
Example~\ref{ex:patch_example9} illustrates a case where negation and graded color terms challenge many of the models.
In the color grid data (Examples~\ref{ex:grid_example3} and \ref{ex:grid_example14}, Table~\ref{tab:examples}), these issues combine with challenges in understanding the basic grid layouts.
For instance, in Example~\ref{ex:grid_example3}, models fail to locate the graded description of a shade of \emph{purple} within the respective grid and seem to simply focus on the locative adjective \emph{right}. Models also fail on complex grid examples like 
Example~\ref{ex:grid_example14}, where descriptions of multiple rows or cells within a grid need to be composed, for a correct resolution of the reference.

\subsection{Discussion}
\label{sec:discussion}
In general, our results do not warrant a conclusive answer to the title question.
While the large variants of the Qwen model achieve close to human performance on the simple color patch task and show promising accuracies in resolving references to more complex color grids, the LLaVA and Janus models do not show robust competencies in either setting. The great difference between the large LLaVA and Qwen models is particularly striking in this regard, suggesting that architectural decisions in MLLMs and fine-tuning protocols can have substantial effects on fundamental capabilities in situated language understanding -- compared to LLaVA,  Qwen uses a different vision encoder and multimodal rotational position encoding, and was subject to multiple pretraining stages (see comparison in Table~\ref{tab:model}, Appendix). Future work should explore such variations more systematically, potentially with smaller LMs, to gain a deeper understanding of the factors that lead to implicit pragmatic competencies in LMs.

In addition, different reasoning skills can be necessary for correctly resolving references. While descriptions often only apply to single items (e.g., \enquote{yellow green} in Figure \ref{fig:patches}), they are more ambiguous in other cases and require reasoning about how the director might have referred to alternative targets (e.g., \enquote{orange} in Example \ref{ex:patch_example11}, see Section \ref{sec:examples}). 
The distinction between these cases is not always clear, which is why our results do not differentiate between different forms of contextual or pragmatic reasoning.

Finally, one advantage that the human matchers had in the reference games is that they can collaborate with the director, e.g., by requesting clarification or more information. In contrast to that, the MLLMs in our investigation are merely ‘overhearers’ \citep{SchoberClark1989} and had to rely on the information specified in the prompt. 
Future work should explore more interactive settings where the agent can negotiate common ground with the director.

\section{Conclusion}
\label{sec:conclusion}

In this paper, we investigated how well MLLMs perform basic reference resolution tasks that require contextualized pragmatic reasoning about color in two abstract visual domains of different complexity. We found that models with sufficient capacity achieve promising results in the simpler domain of color patches, but that even the best-performing models struggle with the more complex structure of color grids.

\section*{Limitations}
\label{sec:limitations}

The study presented in this paper has a number of limitations.
First, we have focused on models whose weights are available and can be run locally. The performance of commercial models such as ChatGPT or Gemini may be different. 
Second, we presented the stimuli as a single image to the MLLMs. Changing the format of the visual input so that each of the three object is provided as an individual image could enable the models to use the full potential of their visual encoders for each stimulus object. This could be particularly beneficial in the color grid domain with more complex objects.
Third, here we did not systematically study the dialogue structure of some of the human interactions.
It would be very interesting to investigate if there are differences between cases where the required information is contained in a single utterances in contrast to cases where it is built up by interlocutors turn-by-turn.
Finally, the order of location labels was fixed (left, middle, right) in our prompts to the models and there is a tendency, at least in some of the models, to prefer the label first mentioned (left). Changing the order in the prompt may mitigate this issue.

\section*{Acknowledgments}

This research has been funded by the \href{https://www.dfg.de/}{Deutsche Forschungsgemeinschaft} (DFG, German Research Foundation) -- \href{https://gepris.dfg.de/gepris/projekt/512393437}{CRC-1646, project no. 512393437}, project \href{https://gepris.dfg.de/gepris/projekt/537416633}{B02}.

\bibliography{unified}

\appendix

\section{Risks and Ethical Considerations}

We do not believe that there are significant risks associated with this work, as we analyze existing models using data with limited scale without contents that might be perceived as hurtful. No ethics review was required.

\section{Prompts}
\label{app:prompts}

Our model prompts consist of three parts: 
    (i) a general task instruction,
    (ii) the formatted utterances for the current item in question, and 
    (iii) a repetition of the set of possible output labels. The prompts are constructed as in the following example from the color grid domain:

\begin{quote}\textit{%
    In this image you can see three color grids. In the following dialogue, the speaker will describe exactly one of the grids. Please indicate to me whether he refers to the left, middle or right grid.}
\end{quote}
\begin{quote}\textit{%
    speaker: first square is brown}
\end{quote}
\begin{quote}\textit{
    Is it the left, middle or right grid?}
\end{quote}

In case a chat dialogue between director (speaker) and matcher (listener) occurs, it is included in part (ii) of the prompt as follows:
\begin{quote}\textit{%
    speaker: CENTER BOX is DULL purple\\
    listener: with bright green on left middle or dull green\\
    speaker: BOTTOM RIGHT CORNER is green}
\end{quote}

\section{Implementation Details}
\label{app:implementation-details}

For our experiments we rely on models from \href{https://huggingface.co/}{huggingface}. In detail, we used the following models:
\begin{itemize}
    \itemsep0em
    \item \href{https://huggingface.co/deepseek-ai/Janus-Pro-1B}{deepseek-ai/Janus-Pro-1B}
    \item \href{https://huggingface.co/deepseek-ai/Janus-Pro-7B}{deepseek-ai/Janus-Pro-7B}
    \item \href{https://huggingface.co/llava-hf/llava-v1.6-vicuna-7b-hf}{llava-hf/llava-v1.6-vicuna-7b-hf}
    \item \href{https://huggingface.co/llava-hf/llava-v1.6-vicuna-13b-hf}{llava-hf/llava-v1.6-vicuna-13b-hf} 
    \item \href{https://huggingface.co/llava-hf/llava-v1.6-34b-hf}{llava-hf/llava-v1.6-34b-hf}
    \item \href{https://huggingface.co/llava-hf/llava-next-72b-hf}{llava-hf/llava-next-72b-hf} (quantized using the \emph{bitsandbytes} library)
    \item \href{https://huggingface.co/Qwen/Qwen2-VL-2B-Instruct}{Qwen/Qwen2-VL-2B-Instruct}
    \item \href{https://huggingface.co/Qwen/Qwen2-VL-7B-Instruct}{Qwen/Qwen2-VL-7B-Instruct}
    \item \href{https://huggingface.co/Qwen/Qwen2-VL-72B-Instruct-AWQ}{Qwen/Qwen2-VL-72B-Instruct-AWQ}
\end{itemize}

To generate responses with our models, we used Python 3.9.20 with the following libraries:
\begin{itemize}
    \itemsep0em
    \item torch (2.5.1)
    \item transformers (4.46.2)
    \item autoawq (0.2.8)
    \item bitsandbytes (0.44.1)
\end{itemize}

We used three NVIDIA RTX A6000 GPUs for inference with the 72b models, two GPUs of the same type for LLaVA 34b and a single GPU of the same type for the remaining models. Depending on model size, generating responses took between \SI{10}{h} and \SI{46}{h} for the color patch data and between \SI{2.5}{h} and \SI{13}{h} for the color grid data.

\begin{figure*}[t]
    \centering
    \begin{subfigure}{.48\textwidth}
        \includegraphics[width=\columnwidth]{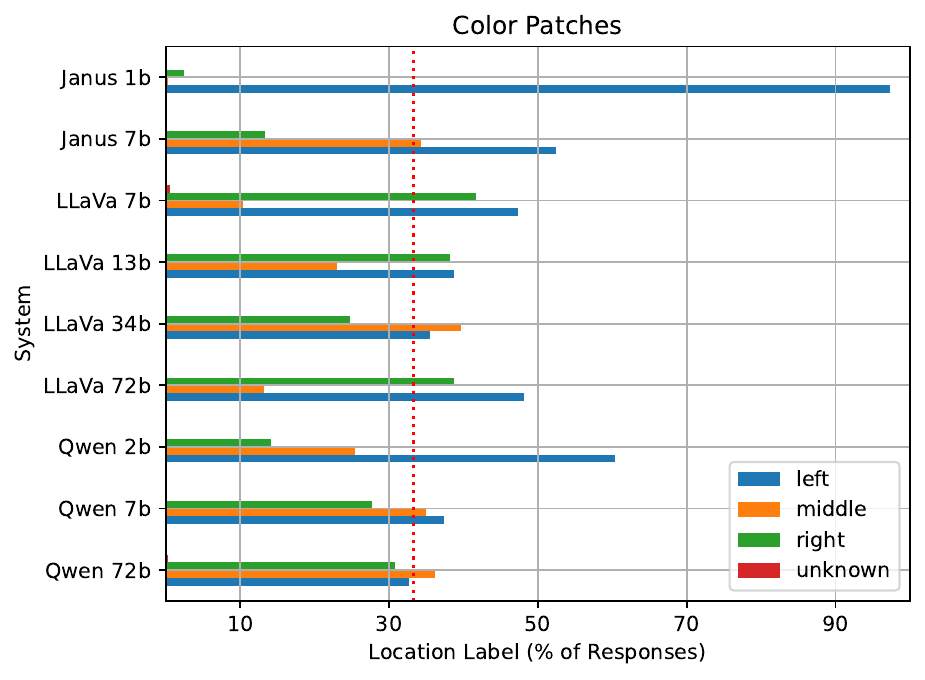}
    \end{subfigure}
    \begin{subfigure}{.48\textwidth}
        \includegraphics[width=\columnwidth]{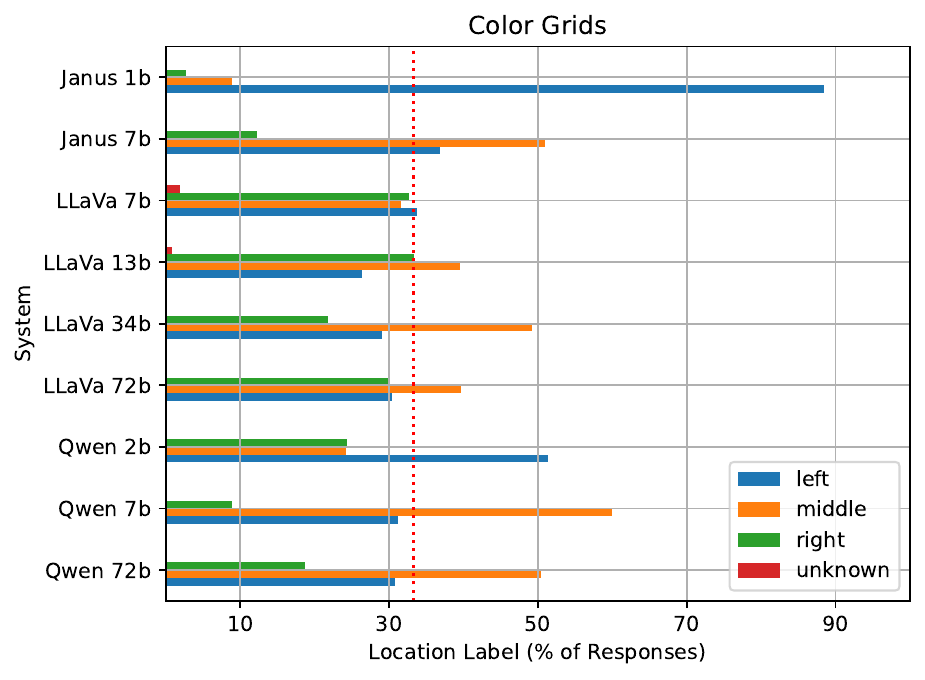}
    \end{subfigure}
    \caption{Location biases in model responses for color patches (left) and color grids (right). The vertical dotted red lines denote the approximately equal distribution of target locations in the data.}
    \label{fig:location_biases}
\end{figure*}

\section{Scientific Artifacts}
\label{app:scientific-artifacts}

In our work, we mainly used scientific artifacts in the form of publicly available datasets and model implementations (MIT, Apache 2.0, and Llama~2 licenses), as well as Python frameworks and modules (cf. Appendix~\ref{app:implementation-details}). 
The color patch dataset can be downloaded from \href{https://cocolab.stanford.edu/datasets/colors.html}{cocolab.stanford.edu}, the color grid dataset is available on \href{https://github.com/forkunited/ltprg}{GitHub} (MIT License). In all cases, we are confident that our work is consistent with their intended use.

Data and code of this study are available (\href{https://www.apache.org/licenses/LICENSE-2.0}{Apache License 2.0}) at: 
\begin{itemize}
    \itemsep0em
    \item \href{https://doi.org/10.5281/zenodo.15553655}{https://doi.org/10.5281/zenodo.15553655}
    \item \href{https://github.com/clause-bielefeld/mllm-listeners}{https://github.com/clause-bielefeld/mllm-listeners}
\end{itemize}

\section{Results for Simplified Dialogues}
\label{app:results-simplified-dialogues}

To test effects of linguistic complexity, we test our models on restricted subsets of the color patch and grid datasets, where we only include dialogues with a total length of $\le 5$ words, out of which $\ge \nicefrac{1}{3}$ have to be included in the following \emph{basic color terms} (cf. \citealt{Berlin1969, Kay1978}):
\emph{black}, \emph{white}, \emph{red}, \emph{green}, \emph{yellow}, \emph{blue}, \emph{brown}, \emph{orange}, \emph{pink}, \emph{purple}, \emph{gray/grey}. See Table~\ref{tab:simplified_results} for detailed results.

\vspace{3mm} 

\section{Location Biases in Model Responses}
\label{app:location-biases}

For color grids we test if location terms in input dialogues introduce biases in model responses by defining subsets with utterances that contain exactly one of the labels \enquote{left}, \enquote{middle} or \enquote{right}. For each of the location labels, we report the proportion of cases (\%) where the model predicts the respective label. The results in Table~\ref{tab:loc_bias} show that all models are affected by location descriptions in input dialogues, albeit to varying degrees. Figure~\ref{fig:location_biases} illustrates model biases for target location predictions with respect to the full datasets.

\balance

\begin{table*}
\small
\begin{tabularx}{\textwidth}{lXXXXXXXXXX}
    \toprule
     &  &  & \multicolumn{4}{c}{\textbf{Color Patches}} & \multicolumn{4}{c}{\textbf{Color Grids}} \\
     \cmidrule(lr){4-7}
     \cmidrule(lr){8-11}
    \textbf{Model} & \textbf{Size (b)} & \textbf{Quant} & \textbf{Total} & \textbf{Far} & \textbf{Split} & \textbf{Close} & \textbf{Total} & \textbf{Far} & \textbf{Split} & \textbf{Close} \\
    \midrule
    \multirow[c]{2}{*}{Janus} & 1 & --- & \textbf{36.9} & \textbf{39.1} & \textbf{36.1} & \textbf{34.8} & 33.1 & 32.7 & \textbf{33.6} & 33.3 \\
     & 7 & --- & \textbf{75.2} & \textbf{89.4} & \textbf{68.9} & \textbf{61.5} & \textbf{41.5} & \textbf{41.7} & \textbf{39.4} & \textbf{43.6} \\
     \midrule
    \multirow[c]{4}{*}{LLaVA} & 7 & --- & \textbf{64.5} & \textbf{74.2} & \textbf{62.2} & \textbf{52.9} & 37.6 & 36.7 & 37.3 & \textbf{39.7} \\
     & 13 & --- & \textbf{62.0} & \textbf{70.1} & \textbf{59.5} & \textbf{52.7} & 36.0 & 35.5 & 35.1 & \textbf{38.1} \\
     & 34 & --- & \textbf{84.9} & \textbf{95.2} & \textbf{80.8} & \textbf{74.6} & 37.5 & 36.9 & 36.9 & \textbf{39.7} \\
     & 72 & 8bit & \textbf{65.2} & \textbf{77.0} & \textbf{60.9} & \textbf{52.6} & \textbf{40.5} & 40.1 & \textbf{40.5} & \textbf{41.1} \\
     \midrule
    \multirow[c]{3}{*}{Qwen} & 2 & --- & \textbf{69.9} & \textbf{83.7} & \textbf{64.4} & \textbf{55.7} & \textbf{41.0} & \textbf{40.8} & \textbf{39.9} & \textbf{42.8} \\
     & 7 & --- & \textbf{87.3} & \textbf{95.9} & \textbf{84.3} & \textbf{77.9} & 44.0 & 45.2 & 41.7 & \textbf{44.7} \\
     & 72 & awq & \textbf{90.1} & \textbf{96.4} & \textbf{88.6} & \textbf{82.5} & 65.9 & 66.9 & 64.9 & \textbf{65.1} \\
     \midrule
    human & --- & --- & \textbf{91.6} & \textbf{97.7} & \textbf{90.7} & \textbf{83.6} & \textbf{94.5} & \textbf{96.8} & \textbf{93.6} & \textbf{90.9} \\
    \bottomrule
\end{tabularx}
\caption{%
    Accuracy scores (\%) for all models and datasets, restricted to items with limited description complexity (max. 5 tokens per item, from which min. \nicefrac{1}{3} have to be basic color terms). Scores which surpass the full dataset results are highlighted in bold. Note that \emph{close} annotations tend to be longer than \emph{far} and \emph{split} annotations, i.e., total scores are skewed towards easier items.}
\label{tab:simplified_results}
\end{table*}

\begin{table*}
\small
\begin{tabularx}{\textwidth}{lXXXXXXXX}
\toprule
     &    &     & \multicolumn{2}{c}{\textbf{Left}} & \multicolumn{2}{c}{\textbf{Middle}} & \multicolumn{2}{c}{\textbf{Right}} \\
    \cmidrule(lr){4-5}
    \cmidrule(lr){6-7}
    \cmidrule(lr){8-9}
\textbf{Model} & \textbf{Size (b)} & \textbf{Quant} & \textbf{Predicted} & \textbf{Accuracy} &  \textbf{Predicted} & \textbf{Accuracy} &  \textbf{Predicted} & \textbf{Accuracy} \\
\midrule
Janus & 1  & --- &         100.0 &     33.1 &            32.2 &       36.2 &           13.1 &      33.7 \\
     & 7  & --- &          94.4 &     34.7 &            95.6 &       34.7 &           53.8 &      41.9 \\
\midrule
LLaVA & 7  & --- &          90.9 &     37.6 &            89.2 &       37.0 &           94.4 &      32.2 \\
     & 13 & --- &          88.7 &     37.6 &            95.4 &       34.8 &           95.8 &      32.5 \\
     & 34 & --- &          92.5 &     34.3 &            99.5 &       33.7 &           90.5 &      32.6 \\
     & 72 & 8bit &          88.1 &     38.4 &            87.1 &       37.2 &           91.1 &      32.5 \\
\midrule
Qwen & 2  & --- &          95.9 &     35.0 &            67.0 &       38.3 &           73.2 &      38.7 \\
     & 7  & --- &          76.1 &     46.4 &            92.8 &       37.2 &           30.7 &      47.4 \\
     & 72 & awq &          47.4 &     71.0 &            69.0 &       55.8 &           35.9 &      74.9 \\
\bottomrule
\end{tabularx}

\caption{%
    Location biases for all models in the \emph{color grid} task. For each of the location labels \enquote{left}, \enquote{middle}, and \enquote{right}, we report the proportion of cases (\%) where the model predicts the respective label, if it is the only location label in the annotations (33.3 \% is the expected result). Accuracy reports the accuracy of predictions (\%) in these cases.}
\label{tab:loc_bias}
\end{table*}


\begin{table*}
\renewcommand{\arraystretch}{1.2}
\small
\begin{tabularx}{\textwidth}{lXXX}
\toprule
    & \textbf{Qwen2-VL} 
    & \textbf{LLaVA-NeXT} 
    & \textbf{Janus-Pro} \\
\midrule
    \textbf{Vision Encoder}
        & ViT and 2D-RoPE 
        & CLIP-ViT-L, higher-res grids 
        & SigLIP-Large-Patch16-384 \\
    \textbf{Cross-modal Connection} 
        & MLP 
        & MLP
        & MLP \\
    \textbf{LLM Backbone} 
        & Qwen2 
        & Vicuna (3b, 7b), NousHer-mes2-Yi (34b), and Llama 3 (72b)
        & Autoregressive Transformer\\
    \textbf{Training} 
        & Pretraining on image-text pairs (ViT), full parameter training and instruction fine-tuning (ViT frozen) 
        & Pretraining cross-modal connection and instruction fine-tuning 
        & Pretraining cross-modal connection, image heads, unified pretraining, and instruction fine-tuning \\
    \textbf{Data} 
        & Extensive and diverse data\-sets 
        & High-quality visual and multimodal data
        & Expanded dataset and synthetic data\\
    \textbf{Special Features} 
        & Naive Dynamic Resolution support and Multimodal Rotary Position Embedding (M-ROPE)
        & Response prompting, dynamic high resolution, and data-efficient
        & Decoupled Visual Encoding\\
\bottomrule 
\end{tabularx}
\caption{%
    Comparison of aspects of the multimodal large language models Qwen2-VL, LLaVA-NeXT, and Janus-Pro used in this study, based on information available in the publications on these models (see Section~\ref{sec:models}).}
\label{tab:model}
\end{table*}

\end{document}